%% file: main.tex
\documentclass[11pt,letterpaper]{article}
\usepackage[utf8]{inputenc}
\usepackage[T1]{fontenc}
\usepackage{lmodern}
\usepackage[margin=1in]{geometry}
\usepackage{amsmath,amssymb,array,booktabs,tabularx,listings,graphicx}
\usepackage{microtype}
\usepackage[numbers,sort&compress]{natbib}
\usepackage{caption}
\usepackage{placeins,float,needspace}
\usepackage{xurl}
\usepackage[hidelinks]{hyperref}
\newcommand{\Description}[1]{}
\author{Zongrui Yang, Li Xintong, Runchen Xu, Zhongsheng Wang,\\
Zhedong Lin, Haoyuan Li, Jiamou Liu}
\title{\textbf{MCRI: A Four-Dimensional Framework for\\Analyzing and Evaluating Agent Skills}}
\date{}
\hypersetup{pdftitle={MCRI: A Four-Dimensional Framework for Analyzing and Evaluating Agent Skills},pdfauthor={Zongrui Yang, Li Xintong, Runchen Xu, Zhongsheng Wang, Zhedong Lin, Haoyuan Li, Jiamou Liu},pdfsubject={Agent skills and pre-execution evaluation}}
\begin{document}
\maketitle
\begin{abstract}
As agents evolve from single-tool systems into modular, composite architectures, skills are becoming an important mechanism for capability development and distribution. However, the academic community lacks a structured framework for systematically analyzing and evaluating skills. Drawing on information gain and behavioral constraint, we propose the four-dimensional MCRI Framework and operationalize it as MCRI-Eval, a large language model–based evaluation method. We evaluate MCRI-Eval using 63,812 public skills from the OpenClaw skill Hub, with 58,275 skill-conditioned model executions across BigCodeBench, BFCL-Fundamental, and Mind2Web. MCRI-Eval scores are positively associated with community popularity signals and achieve the highest downstream ranking agreement among the evaluated methods. MCRI-Eval also improves top-1 skill selection across all three benchmarks: compared with the strongest baseline on each benchmark, the skills selected by MCRI-Eval advance by 17.7, 22.8, and 19.6 percentile points in downstream performance rank on BigCodeBench, BFCL-Fundamental, and Mind2Web, respectively. These results indicate that MCRI-Eval provides a useful pre-execution signal for prioritizing promising skills before costly execution-based evaluation.

\end{abstract}
\input{main_body}

\FloatBarrier
\bibliographystyle{plainnat}
{\small\raggedright
\bibliography{main_references,supplement_references}
}
\clearpage
\appendix
\input{supplement_body}

\end{document}

%% file: main_body.tex
\section{Introduction}

AI agents are increasingly integrated into contemporary workflows. Rather than functioning as isolated assistive modules, they are evolving into composite systems built from reusable capability modules. Claude Code provides a representative example. On October 16, 2025, Anthropic introduced Agent skills~\cite{anthropic2025skills} and subsequently packaged several built-in capabilities as skills. Major vendors soon followed~\cite{openai_codex_agent_skills,google2026agent_skills,cursor2026agent_skills}, constructing modular capability ecosystems around MCP, skill Hubs, and skill installation interfaces. At the time of writing, LobeHub's ``skills Marketplace for Every Agent'' has collected more than 290,000 skills~\cite{lobehub2026skills_marketplace}.

These developments suggest that skills are moving beyond auxiliary configurations in engineering practice and becoming important building blocks in agent capability architectures. Despite their growing practical importance, however, neither the academic nor the engineering community has established a structured perspective for systematically analyzing, organizing, and evaluating skills. Why are skills effective? What characteristics make a skill more likely to gain widespread adoption? Without such a perspective, it remains unclear how skills should be optimized or how users can identify promising candidates within increasingly large repositories.

Accordingly, we formulate the following research questions:

\begin{itemize}
\item \textbf{RQ1:} Through which functional mechanisms and structural dimensions can skills be systematically characterized?
\item \textbf{RQ2:} Do content-based scores derived from these dimensions capture meaningful signals associated with real-world ecosystem outcomes?
\item \textbf{RQ3:} Can these scores predict the relative performance of skills on downstream tasks and support the selection of high-performing skills?
\end{itemize}

\begin{figure}[t]
\centering
\includegraphics[width=0.78\textwidth]{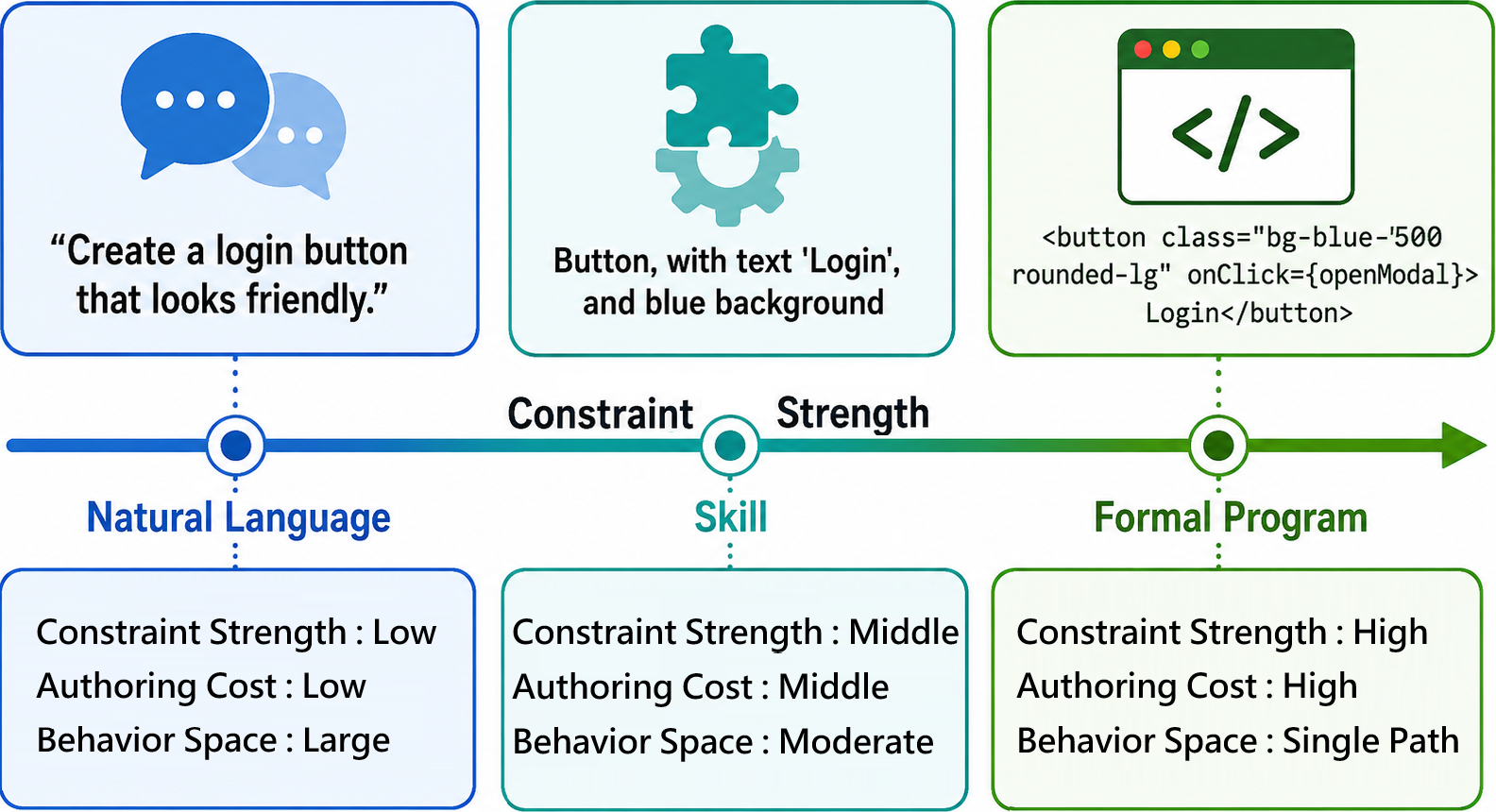}
\caption{Constraint strength spectrum from natural language to skills and formal programs.}
\label{fig:constraint-spectrum}
\Description{A spectrum showing increasing behavioral constraint from natural language through skills to formal programs.}
\end{figure}

To answer these questions, we propose MCRI, an analytical framework for characterizing skills. As illustrated in Figure~\ref{fig:constraint-spectrum}, we conceptualize a skill as an intermediate structure situated between natural language and programming languages. We characterize how skills influence large language model behavior through information gain and behavioral constraint.

More specifically, we argue that the practical functions of a skill are primarily reflected in two mechanisms. First, a skill introduces information external to the model, including task context, domain knowledge, tool descriptions, examples, and external resources. Second, a skill reduces the instability of model behavior through task decomposition, procedural specifications, interface constraints, boundary conditions, and exception-handling mechanisms. In other words, a high-quality skill does not merely provide the model with more text. Instead, through effective information gain and appropriate behavioral constraints, it enables the model to complete a particular family of tasks in a more stable, controllable, and reusable manner.

We evaluate MCRI-Eval using more than 60,000 skills and their platform metadata from the OpenClaw skill Hub. The first experiment examines associations with stars, downloads, and installs across scoring models, sample thresholds, and dimensional ablations. The second compares MCRI-Eval with actual performance across multiple downstream domains, evaluating both full-sample ranking agreement and top-$k$ skill selection.

MCRI-Eval scores are positively associated with ecosystem signals and achieve the highest downstream ranking agreement among the evaluated methods. For top-1 selection, we measure where the skill selected by each method ranks in actual downstream performance within the corresponding benchmark-specific candidate pool. On BigCodeBench~\cite{BCB}, the strongest baseline selects a skill at the 60.4th percentile, whereas MCRI-Eval selects one at the 78.0th percentile, an improvement of 17.7 percentile points. The corresponding percentiles are 51.8 versus 74.6 on BFCL-Fundamental~\cite{BFCL} and 79.7 versus 99.3 on Mind2Web~\cite{MINDWEB}, representing improvements of 22.8 and 19.6 percentile points, respectively.

The appendices provide additional experiments, dataset descriptions, and the complete MCRI-Eval scoring rubric and prompts. Fully reproducing all experiments is estimated to require approximately 0.4 billion tokens. The datasets, code packages, experiment logs, outputs from LLM calls, downstream task analysis results, and test code are not included in this preprint's source package. Appendix~\ref{sec:provenance} describes the availability of these resources.

\section{Related Work}

\subsection{Large Language Models and Agents Perspective}

Large language model evaluation typically uses static benchmarks to measure knowledge and reasoning encoded in model parameters~\cite{related_work_1}. Agent evaluation instead focuses on interactive behavior, including end-to-end task completion, multi-turn interaction, and behavioral consistency~\cite{related_work_3}, as exemplified by AgentBench~\cite{related_work_23} and AgentBoard~\cite{related_work_27}.

Neither paradigm directly evaluates skill content. LLM benchmarks target intrinsic model capability, whereas agent outcomes depend on the host architecture, planning strategy, and tool orchestration. skills are reusable, pluggable artifacts within these systems and therefore constitute a distinct evaluation object.

\subsection{Prompt Engineering Perspective}

Skills are often compared with prompts because both use natural-language instructions and structured specifications. Prompt evaluation includes frameworks such as PromptRobust~\cite{related_work_29}, but usually assumes a single, flat input for a specific task instance. skills instead package reusable capabilities with tool invocation, file-processing rules, API specifications, multi-file resources, preconditions, and exception branches.

These structural and operational differences prevent prompt-oriented evaluation methods from being transferred directly to skills.

\subsection{Software Engineering Perspective}

Skills also resemble third-party software libraries. Product quality models such as ISO/IEC 25010 provide established software-quality dimensions~\cite{related_work_13}, but their evaluation objects and usage assumptions differ from those of natural-language capability modules embedded in agents. A software perspective is therefore informative but insufficient on its own.

\begin{figure*}[t]
    \centering
    \includegraphics[width=\textwidth]{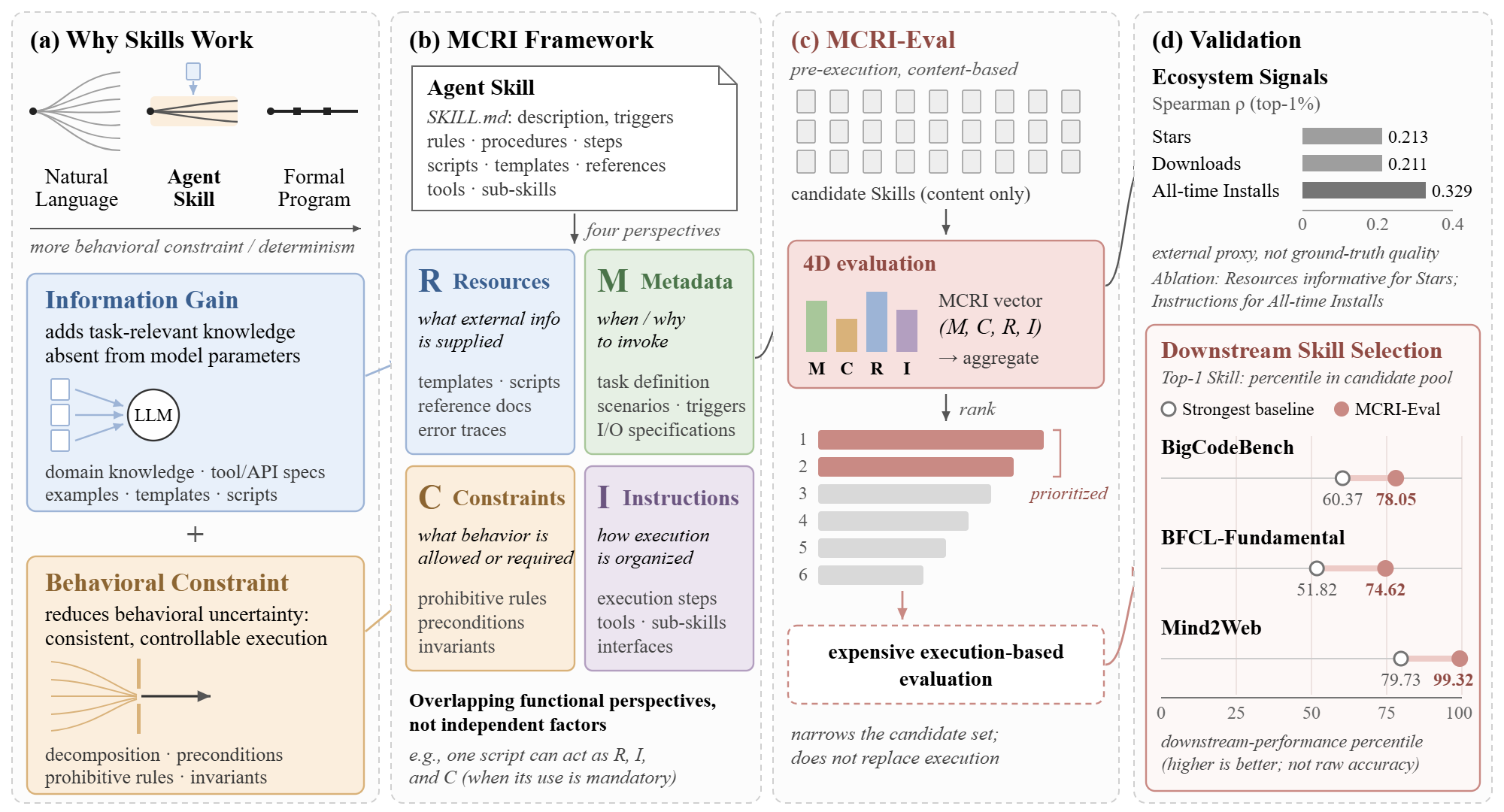}
    \caption{The MCRI framework.}
    \label{fig:mcri}
\end{figure*}

\section{How Do Skills Work?}

\subsection{Where Does a Skill Sit Between Prompts and Code?}

Instructions for computational systems can be arranged by behavioral constraint. As shown in Figure~\ref{fig:constraint-spectrum}, natural-language conversation and formal programs occupy the two ends of this spectrum, while skills lie between them.

\textbf{Natural language conversation.} A request such as ``help me organize this data'' leaves the procedure, output format, and execution steps unspecified. The model must infer the user's intent and choose a strategy, preserving flexibility but producing high behavioral uncertainty.

\textbf{Programming languages.} A program explicitly specifies inputs, operations, branches, and outputs, ideally compressing behavior into a deterministic path. This determinism requires greater authoring effort and limits adaptation beyond the specification.

\textbf{Skills.} A skill selectively constrains behaviors that should be determinate while leaving other choices to model reasoning. Like a software requirements document, it specifies what should be done and under which conditions without prescribing every implementation step.

\subsection{How Skills Work: Constraint and Information Injection}

This intermediate position gives rise to two complementary mechanisms.

\subsubsection{Information Gain}

Model parameters cannot contain all task-specific information, such as an internal API format, a project's directory conventions, or recent library changes. Stronger reasoning alone cannot recover information absent from the model's parameters.

A PPTX-generation skill, for example, may provide concrete \texttt{python-pptx} usage and template-manipulation code. We call this mechanism \textbf{information gain}: the skill expands the model's effective knowledge with task-relevant external information, making the appropriate behavioral path available.

\subsubsection{Behavioral Constraint}

Even with sufficient information, 
probabilistic generation may produce different strategies, output formats, or task decompositions across runs. Several paths may be reasonable, but only some satisfy the requirements of a particular setting.

For a CSV-analysis request, for example, the model may begin with missing-value handling, visualization, statistical testing, or clarification. None is necessarily wrong, but such variation reduces reusability.

A skill that prescribes preprocessing, analysis, and a report format compresses the model's permissible behavior space. We call this mechanism \textbf{behavioral constraint}: it increases the consistency with which a capable model selects the target path across repeated invocations.

\subsection{MCRI Framework: An Analytical Framework for Skills}

Based on the two mechanisms identified in Section~3.2, we propose the MCRI Framework, represented as the following four-tuple:
\begin{equation}
MCRI(\mathrm{skill}) =
\bigl(
M(\mathrm{skill}),
C(\mathrm{skill}),
R(\mathrm{skill}),
I(\mathrm{skill})
\bigr).
\end{equation}

\noindent\textbf{$M$ (Metadata): Identification Layer.}
Metadata describes the task definition, functional role, applicable scenarios, triggering conditions, and input--output specifications associated with a skill. It informs the host agent not only of ``what this skill is designed to do,'' but also of ``under what circumstances it should be selected and invoked.''

\noindent\textbf{$C$ (Constraints): Constraint Layer.}
Constraints correspond to the behavioral constraint mechanism discussed in Section~3.2. This dimension captures the parts of a skill that explicitly restrict the model's space of permissible behaviors, including prohibitive rules, preconditions, invariants, and prescribed handling strategies.

\noindent\textbf{$R$ (Resources): Information Layer.}
Resources correspond to the information-gain mechanism discussed in Section~3.2. This dimension comprises information injected into the model from outside its parameters, including template files, configuration scripts, example data, reference documents, and error traces accumulated by others during prior executions.

\noindent\textbf{$I$ (Instructions): Interface Layer.}
Instructions specify the internal scripts, external tools, and sub-skills that may be invoked during execution, as well as the concrete means by which the skill interacts with its environment. This dimension includes interface specifications, the ordering of execution steps, and the logic used to decompose a task into subtasks.

\begin{figure}[t]
    \centering
    \includegraphics[width=0.78\textwidth]{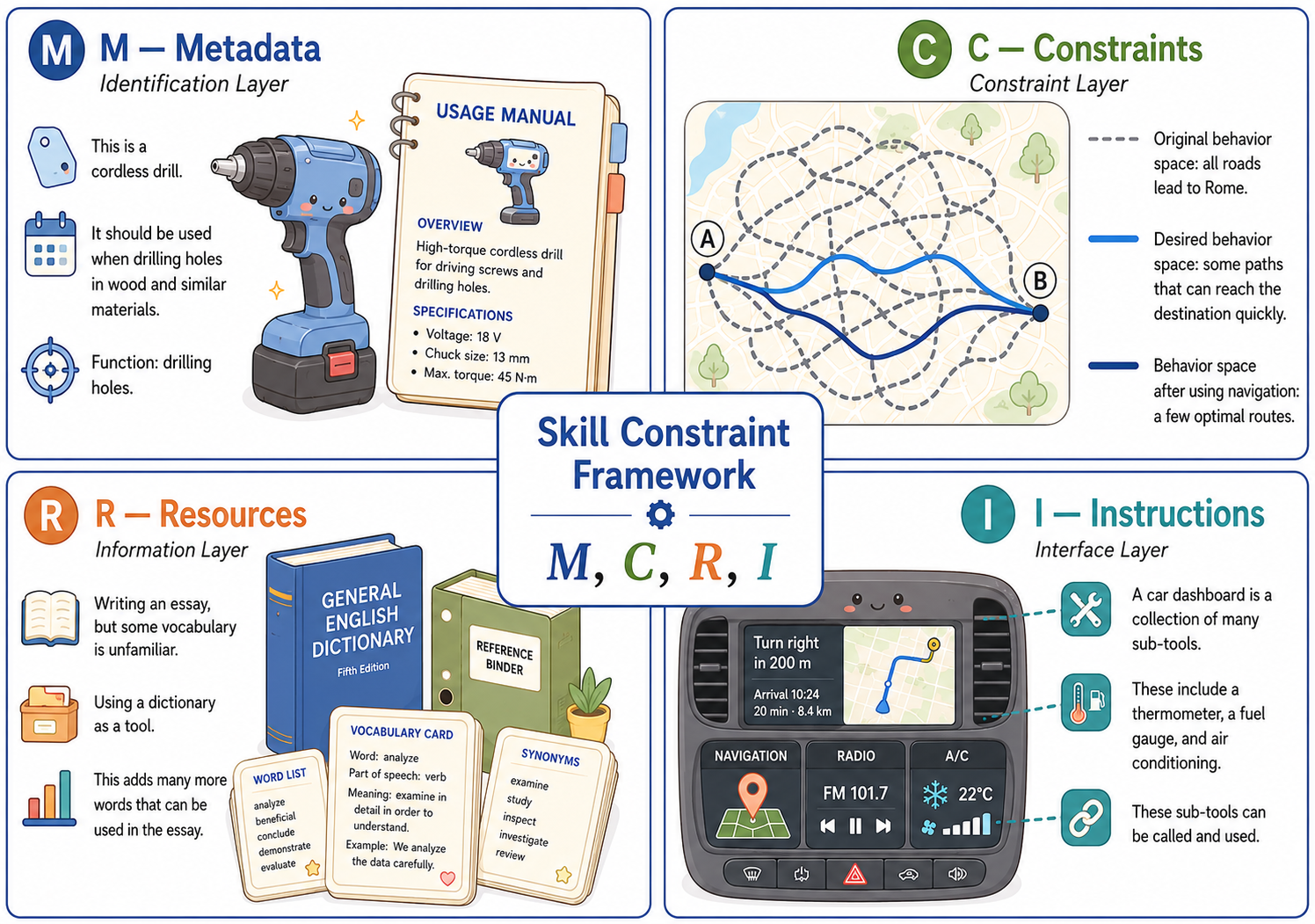}
    \caption{Example of the MCRI Framework. }
    \label{fig:skill-constraint-framework}
\Description{An illustration of the Metadata, Constraints, Resources, and Instructions dimensions in the MCRI framework.}
\end{figure}

MCRI defines four functional perspectives rather than mutually exclusive components or statistically independent latent factors. Their distinction lies in the question each perspective asks: Metadata concerns whether the host can identify when and why to invoke a skill; Constraints concern how permissible behavior is restricted; Resources concern what external information is supplied; and Instructions concern how execution is organized and connected to the environment. The same artifact may therefore affect several dimensions for different functional reasons. A script, for example, can supply implementation knowledge as a Resource, expose a callable interface as an Instruction, and impose a Constraint when its use is mandatory.

Overlap does not make the dimensions interchangeable. MCRI-Eval scores each function with a separate rubric, and the ablations in Table~\ref{tab:ablation-stars-installs} show differentiated effects: removing Resources primarily weakens the association with stars, whereas removing Instructions most strongly affects all-time installs. These results support functional differentiation without requiring orthogonality among the four dimensions.

Downstream execution remains the most direct evidence of effectiveness, but it is available only after reasoning and tool-calling costs have been incurred. MCRI instead provides a structured feature space for pre-execution analysis of identifiability, constraint sufficiency, resource relevance, and instruction executability. Its scores are proxies rather than substitutes for downstream validation; their purpose is to narrow the candidate set and focus execution-based evaluation on more promising skills.

\section{How Well Do Skills Work?}
\label{sec:skill-effectiveness}

\subsection{From MCRI to MCRI-Eval}
\label{subsec:mcri-eval}

We extend the MCRI Framework from a descriptive framework into an
evaluation heuristic. Each layer corresponds to an evaluation dimension that
measures how effectively the layer fulfills its intended function.
Specifically, \emph{Metadata} corresponds to task-description clarity,
\emph{Constraints} to behavioral-constraint sufficiency, \emph{Resources} to
the effectiveness of information gain, and \emph{Instructions} to the
completeness and standardization of instructions and interfaces. The complete
evaluation prompt contains the definition of the MCRI Framework and the scoring
rubric. The full prompt is provided in Appendix~\ref{sec:rubric}.

\subsection{Community-Signal Identification}
\label{subsec:community-signals}

\subsubsection{Task Definition}

Given a skill $S_i$, an LLM produces a four-dimensional score vector
$\mathbf{L}_i \in \mathbb{R}^{1\times4}$ according to the MCRI framework. On
the training set, we fit a linear model from $\mathbf{L}_i$ to a target
community metric $f_i$, yielding an aggregate LLM-based prediction $g_i$. We
then apply the fitted model to the test set and compare the within-set rankings
induced by $\{f_i\}$ and $\{g_i\}$ to evaluate correlation and ranking
accuracy.

\paragraph{Correlation identification.}
Task T1 computes the Spearman correlation coefficients and significance levels
between the four MCRI dimensions and several community metrics.

\paragraph{Extreme-class identification.}
Task T2 treats the bottom 15\% of skills as the low-signal group and the top
15\% as the high-signal group, and then evaluates binary classification
performance.

\subsubsection{Experimental Setup}

\paragraph{Dataset.}
We collected 63,812 public skills from the OpenClaw skill Hub, including the
skill contents and observable platform signals. Raw download and installation
counts are affected by how long a skill has been available and by its exposure
on the platform. Moreover, the dataset exhibits a pronounced long-tail
distribution, and most community metrics are zero. To reduce the resulting interference in the correlation analysis, we compute the average daily
downloads for every sample and conduct the analysis only on the highest-ranked
skills. We randomly split all samples into equally sized training and test sets
using a fixed random seed (\texttt{random\_state=42}) for reproducibility. The
training set is used for fitting, whereas the test set is reserved exclusively
for final evaluation.

\paragraph{Scoring models.}
We use models from the MiMo family (MiMo-v2.5 and MiMo-v2.5-Pro) and the GPT
family (GPT-5.4-nano and GPT-5-mini) as scoring models. Each model assigns 
four-dimensional scores to every skill according to the proposed evaluation
heuristic. The GPT models are developed by OpenAI, whereas the MiMo models are
developed by Xiaomi.

\subsubsection{Results}

This section examines the central research question: can four-dimensional LLM
assessments of skill text alone identify skills that are more popular in a real
ecosystem? We evaluate four scoring models under three filtering
thresholds---top 3\%, top 1\%, and top 0.5\%---to determine how predictive
strength changes as the analysis concentrates on increasingly prominent
samples.

\paragraph{Overall trend across thresholds.}
Table~\ref{tab:best-correlation-thresholds} reports, for each community signal
and filtering threshold, the strongest Spearman correlation obtained by any of
the four scoring models. Correlations increase monotonically as the threshold
becomes more selective. This pattern suggests that differences among tail samples contain substantial noise, making community metrics less reliable as proxies for quality in that region.

\begin{table*}[t]
\caption{Strongest Spearman correlations at different filtering thresholds.
  The corresponding scoring model is shown in parentheses.}
\label{tab:best-correlation-thresholds}
  \centering
  \footnotesize
  \setlength{\tabcolsep}{4pt}

  \begin{tabular}{lcc cc cc}
    \toprule
    \textbf{Signal}
      & \textbf{Top 3\% ($N=1{,}823$)}
      & \textbf{Sig.}
      & \textbf{Top 1\% ($N=606$)}
      & \textbf{Sig.}
      & \textbf{Top 0.5\% ($N=305$)}
      & \textbf{Sig.} \\
    \midrule
    Stars
      & 0.166 (GPT-5.4-nano)
      & ***
      & 0.213 (GPT-5.4-nano)
      & ***
      & 0.302 (MiMo-v2.5)
      & *** \\
    Downloads
      & 0.125 (GPT-5-mini)
      & ***
      & 0.211 (GPT-5.4-nano)
      & ***
      & 0.293 (MiMo-v2.5)
      & *** \\
    All-time installs
      & 0.197 (GPT-5.4-nano)
      & ***
      & 0.329 (GPT-5.4-nano)
      & ***
      & 0.393 (MiMo-v2.5)
      & *** \\
    \bottomrule
  \end{tabular}

\end{table*}

\paragraph{Cross-model consistency.}
Table~\ref{tab:top1-model-correlations} further reports correlations for all
four scoring models at the top-1\% threshold. Most model--signal pairs are
statistically significant, although a small number of weaker or nonsignificant
cases remain. The cross-model spread is moderate, and no choice of scoring
model reverses the overall conclusion. This consistency suggests that the
observed relationships are not driven solely by an idiosyncratic bias of a
particular scoring model.

\begin{table*}[t]
\caption{Spearman correlations at the top-1\% threshold across four
  scoring models. Statistical significance is denoted by
  $^{*}p<0.05$, $^{**}p<0.01$, and $^{***}p<0.001$;
  $\mathrm{ns}$ denotes a nonsignificant result.}
\label{tab:top1-model-correlations}
  \centering
  \setlength{\tabcolsep}{5pt}

  \begin{tabular}{@{}lcccc@{}}
    \toprule
    \textbf{Signal}
      & \textbf{MiMov2.5Pro}
      & \textbf{MiMov2.5}
      & \textbf{GPT5.4-nano}
      & \textbf{GPT5-mini} \\
    \midrule
    Stars
      & $+0.203^{***}$
      & $+0.152^{**}$
      & $+0.213^{***}$
      & $+0.134^{*}$ \\
    Downloads
      & $+0.165^{**}$
      & $+0.183^{**}$
      & $+0.211^{***}$
      & $+0.082^{\mathrm{ns}}$ \\
    Comments
      & $+0.156^{**}$
      & $+0.152^{**}$
      & $+0.270^{***}$
      & $+0.208^{***}$ \\
    All-time installs
      & $+0.280^{***}$
      & $+0.267^{***}$
      & $+0.329^{***}$
      & $+0.226^{***}$ \\
    \bottomrule
  \end{tabular}

\end{table*}

\paragraph{Dimension-level ablation.}
We remove one dimension at a time, refit $g_i$, and compare correlations on
the top-1\% samples. Table~\ref{tab:ablation-stars-installs} shows distinct
effects. Removing Resources makes the correlation with stars nonsignificant for
three scoring models, reducing $\rho=0.13$--$0.21$ to $\rho=0.04$--$0.06$.
Removing Instructions produces the largest decline for all-time installs,
whereas removing Constraints yields smaller, broadly consistent decreases and
removing Metadata has limited effect. The dimensions are therefore functionally related but not interchangeable: information-bearing resources are most informative for user attention, while executable instructions are most informative for installation behavior.

\begin{table*}[t]
\caption{Dimension-level ablation results for stars and all-time installs
  across four scoring models.}
\label{tab:ablation-stars-installs}
  \centering
  \footnotesize
  \setlength{\tabcolsep}{5pt}

  \setlength{\tabcolsep}{3pt}
\begin{minipage}[t]{0.49\linewidth}\centering
\textbf{(a) Stars}\par\smallskip
\begin{tabular}{lcccc}
    \toprule
    \textbf{Variant}
      & \textbf{\shortstack{MiMo\\v2.5Pro}}
      & \textbf{\shortstack{MiMo\\v2.5}}
      & \textbf{\shortstack{GPT5.4-\\nano}}
      & \textbf{\shortstack{GPT5-\\mini}} \\
    \midrule
    Full
      & $+0.203^{***}$
      & $+0.152^{**}$
      & $+0.213^{***}$
      & $+0.134^{*}$ \\
    w/o $M$
      & $+0.199^{***}$
      & $+0.169^{**}$
      & $+0.182^{**}$
      & $+0.124^{*}$ \\
    w/o $R$
      & $+0.036^{\mathrm{ns}}$
      & $+0.057^{\mathrm{ns}}$
      & $+0.172^{**}$
      & $+0.039^{\mathrm{ns}}$ \\
    w/o $C$
      & $+0.169^{**}$
      & $+0.140^{*}$
      & $+0.172^{**}$
      & $+0.096^{\mathrm{ns}}$ \\
    w/o $I$
      & $+0.192^{***}$
      & $+0.098^{\mathrm{ns}}$
      & $+0.162^{**}$
      & $+0.127^{*}$ \\
    \bottomrule
  \end{tabular}
\end{minipage}\hfill\begin{minipage}[t]{0.49\linewidth}\centering
\textbf{(b) All-Time Installs}\par\smallskip
\begin{tabular}{lcccc}
    \toprule
    \textbf{Variant}
      & \textbf{\shortstack{MiMo\\v2.5Pro}}
      & \textbf{\shortstack{MiMo\\v2.5}}
      & \textbf{\shortstack{GPT5.4-\\nano}}
      & \textbf{\shortstack{GPT5-\\mini}} \\
    \midrule
    Full
      & $+0.280^{***}$
      & $+0.267^{***}$
      & $+0.329^{***}$
      & $+0.226^{***}$ \\
    w/o $M$
      & $+0.277^{***}$
      & $+0.268^{***}$
      & $+0.288^{***}$
      & $+0.234^{***}$ \\
    w/o $R$
      & $+0.274^{***}$
      & $+0.265^{***}$
      & $+0.334^{***}$
      & $+0.181^{**}$ \\
    w/o $C$
      & $+0.254^{***}$
      & $+0.234^{***}$
      & $+0.316^{***}$
      & $+0.198^{***}$ \\
    w/o $I$
      & $+0.155^{**}$
      & $+0.087^{\mathrm{ns}}$
      & $+0.165^{**}$
      & $+0.112^{\mathrm{ns}}$ \\
    \bottomrule
  \end{tabular}
\end{minipage}

\end{table*}

\paragraph{Extreme-class identification.}
We further investigate whether the scores can distinguish the most popular
subset within the collection of top-ranked skills. Within the top-1\% samples,
we use the top 15\% and bottom 15\% of each community signal to define high-
and low-signal groups. We then use the fitted function $g_i$, based on the
four-dimensional scores, for binary classification. As shown in
Table~\ref{tab:extreme-classification}, performance across the three signals
follows the same ordering as the correlation strengths reported in
Tables~\ref{tab:best-correlation-thresholds}
and~\ref{tab:top1-model-correlations}.

\begin{table*}[t]
\caption{Extreme-class identification within the top-1\% samples. High- and
  low-signal groups correspond to the top and bottom 15\%, respectively.}
\label{tab:extreme-classification}
  \centering
  \footnotesize
  \setlength{\tabcolsep}{3pt}

  \begin{minipage}[t]{0.48\textwidth}
    \centering
    \textbf{(a) All-Time Installs}\par\vspace{2pt}
    \begin{tabular}{lccc}
      \toprule
      \textbf{Model}
        & \textbf{AUC}
        & \textbf{Accuracy}
        & \textbf{F1} \\
      \midrule
      MiMo-v2.5-Pro & 0.758 & 70.5\% & 0.644 \\
      MiMo-v2.5     & 0.719 & 62.9\% & 0.581 \\
      GPT-5.4-nano  & 0.819 & 76.2\% & 0.747 \\
      GPT-5-mini    & 0.716 & 68.6\% & 0.593 \\
      Average       & 0.753 & 69.6\% & 0.641 \\
      \bottomrule
    \end{tabular}
  \end{minipage}
  \hfill
  \begin{minipage}[t]{0.48\textwidth}
    \centering
    \textbf{(b) Downloads}\par\vspace{2pt}
    \begin{tabular}{lccc}
      \toprule
      \textbf{Model}
        & \textbf{AUC}
        & \textbf{Accuracy}
        & \textbf{F1} \\
      \midrule
      MiMo-v2.5-Pro & 0.734 & 64.6\% & 0.623 \\
      MiMo-v2.5     & 0.746 & 67.1\% & 0.658 \\
      GPT-5.4-nano  & 0.756 & 69.5\% & 0.706 \\
      GPT-5-mini    & 0.676 & 64.6\% & 0.603 \\
      Average       & 0.728 & 66.4\% & 0.648 \\
      \bottomrule
    \end{tabular}
  \end{minipage}
  \par\medskip
  \begin{minipage}[t]{0.48\textwidth}
    \centering
    \textbf{(c) Stars}\par\vspace{2pt}
    \begin{tabular}{lccc}
      \toprule
      \textbf{Model}
        & \textbf{AUC}
        & \textbf{Accuracy}
        & \textbf{F1} \\
      \midrule
      MiMo-v2.5-Pro & 0.653 & 60.2\% & 0.545 \\
      MiMo-v2.5     & 0.662 & 59.3\% & 0.540 \\
      GPT-5.4-nano  & 0.700 & 67.3\% & 0.648 \\
      GPT-5-mini    & 0.589 & 54.9\% & 0.427 \\
      Average       & 0.651 & \textbf{60.4\%} & 0.540 \\
      \bottomrule
    \end{tabular}
  \end{minipage}

\end{table*}

\paragraph{Conclusion.}
These results show that the four-dimensional scores derived from the MCRI Framework are significantly and positively associated with skill popularity across multiple independent community signals, scoring models, and filtering thresholds. Among the evaluated signals, all-time installs yield the most stable correlations and classification performance.

\subsection{Correlation with Downstream Task Performance}
\label{subsec:downstream-performance}

\subsubsection{Task Definition}
\label{subsubsec:downstream-task-definition}

For each benchmark, let $U=\{S_i\}_{i=1}^{N}$ denote the set of skills matched to that benchmark. For BigCodeBench and BFCL-Fundamental, we randomly sample $n_B=50$ subtasks to construct the downstream task set $T=\{T_j\}_{j=1}^{n_B}$. Each task is executed independently three times to reduce the effect of generation randomness. For Mind2Web, we sample 50 task chains that jointly contain 345 annotated steps, each of which is executed once.

For each skill $S_i$, we inject the skill into the model execution environment
and use llm to complete the corresponding downstream tasks. All outputs
are evaluated using the official protocol of the corresponding benchmark. For
BigCodeBench and BFCL-Fundamental, we first average the three runs for each task
and then average over the 50 tasks. For Mind2Web, we compute Step Success Rate
over the 345 steps. This yields the downstream performance $p_i$ for each skill.
With 83, 133, and 75 matched skills for BigCodeBench, BFCL-Fundamental, and
Mind2Web, respectively, the protocol produces 58,275 skill-conditioned model
execution calls.

In addition, MCRI-Eval evaluates each skill $S_i$ along four dimensions and
produces the score vector $\mathbf{MCRI}_i=(M_i,C_i,R_i,I_i)$, where $M_i$, $C_i$, $R_i$, and $I_i$ denote the scores for Metadata, Constraints, Resources, and Instructions, respectively.

We compared the following three skill evaluation methods. To ensure consistent evaluation conditions, all methods used GPT-5.4-nano as the scoring model, with identical system prompts and configurations. During downstream task generation, to more thoroughly assess the performance of different models and minimize potential bias arising from the same model being involved in both task generation and evaluation, only BFCL used GPT-5.4-nano to generate tasks, whereas BigCodeBench and Mind2Web both used DeepSeek-V4-Flash.

\textbf{Baseline.}
The scoring model evaluates the overall suitability of a skill using only
its name, summary, and downstream task description. The resulting score is
denoted by $b_i$.

\textbf{Full-text Holistic Baseline.}
In addition to the inputs used by the Baseline, the scoring model is given
the complete skill text and performs a holistic evaluation. The resulting
score is denoted by $bf_i$.

\textbf{MCRI-Eval.}
The task-facing inputs are the same as those used by the Baseline. The
scoring model is additionally provided with the definitions of the four
MCRI dimensions, the corresponding scoring rubrics, and the
four-dimensional evaluation result $\mathbf{MCRI}_i$. It then produces an
aggregate score denoted by $mcri_i$.

All evaluation methods use the same scoring model. The downstream task experiments also use the same generation model, system prompt, tool configuration, and inference budget, thereby controlling, as far as possible, for factors unrelated to the evaluation method.

\paragraph{Correlation analysis.}
We compute full-sample ranking agreement between the downstream performance
sequence $\{p_i\}$ and each of the three evaluation score sequences
$\{b_i\}$, $\{bf_i\}$, and $\{mcri_i\}$. We report the concordance index
(C-index), Spearman's rank correlation coefficient $\rho$, and its associated
significance level $p$.

\paragraph{Skill selection.}
We rank skills in descending order according to $\{b_i\}$, $\{bf_i\}$, and
$\{mcri_i\}$, respectively. For each method, we select the top-$k$ skills and
calculate their average downstream task performance.

\subsubsection{Experimental Setup}
\label{subsubsec:downstream-setup}

\paragraph{Task suites.}
We select three widely used agent benchmarks as downstream task suites:
BigCodeBench, BFCL-Fundamental, and Mind2Web. These benchmarks cover code
generation, function calling, and multi-step web interaction,
respectively.

\textbf{BigCodeBench.}
Downstream performance is measured by code execution pass rate. After a
skill is injected, the model generates a complete function implementation
for each problem, which is then evaluated using the official unit test suite. The final metric is the average test pass rate over 50 problems.

\textbf{BFCL-Fundamental.}
Downstream performance is measured by function-calling success rate. After
a skill is injected, the model returns the expected function call for each
task, which is compared against the officially annotated reference call. The final metric is the average test pass rate over 50 problems.

\textbf{Mind2Web.}
We follow the official task format and implementations of Element Accuracy, Operation F1, and Step Success Rate. The 50 task chains contain 345 annotated steps in total. The downstream performance reported in this study is Step Success Rate over these steps.

\paragraph{Dataset.}
We continue to use the OpenClaw skill Hub dataset introduced in the preceding section. We
first select the top 3\% of skills according to average daily downloads,
resulting in $N=1{,}823$ samples. We then use an LLM to classify all samples by their intended purposes and map each skill to the most applicable
benchmark according to its declared functionality. skills that cannot be
clearly matched to any of the three benchmarks are labeled as
\texttt{unmatched}.

\paragraph{Evaluation protocol.}
All benchmarks are evaluated following their official evaluation settings. Since this work primarily focuses on the ability of different methods to identify and select high-performing skills, and because the original downstream-task performance of most skills is relatively concentrated, comparisons based solely on absolute performance differences may not adequately reflect their relative quality. Therefore, in addition to reporting the original downstream-task performance, we also report the percentile score of each selected skill within its corresponding candidate set.

Specifically, for a benchmark containing (N) candidate skills, we rank all candidates in descending order according to their original downstream-task performance. Let $\mathrm{rank}$ denote the ranking position of the selected skill, where $\mathrm{rank}=1$ indicates the best-performing candidate. When multiple skills achieve identical downstream-task performance, their tied rank is defined as the average of the ranking positions they occupy. The percentile score is then computed as $100 \cdot \frac{N-\mathrm{rank}}{N-1}.$

This score ranges from $0$ to $100$, where $100$ indicates the best performance within the candidate set and $0$ indicates the worst. Under the average-rank convention, skills with identical downstream-task performance receive the same percentile score.

\subsubsection{Experimental Results}
\label{subsubsec:downstream-results}

\begin{table}[t]
\caption{Downstream evaluation results on BFCL-Fundamental.
  Top-$k$ results report downstream performance (Perf.) and the
  corresponding percentile score within the candidate set.}
\label{tab:bfcl-results}
  \centering
  \footnotesize
  \setlength{\tabcolsep}{3pt}

  \textbf{(a) Ranking agreement}\par\vspace{2pt}

  \begin{tabular}{@{}lccc@{}}
    \toprule
    \textbf{Method}
      & \textbf{C-index}
      & \textbf{Spearman $\rho$}
      & \textbf{$p$-value} \\
    \midrule
    Baseline
      & 0.5101
      & 0.0248
      & 0.7765 \\
    Full-text
      & 0.5522
      & 0.1496
      & 0.0856 \\
    MCRI-Eval
      & \textbf{0.5777}
      & \textbf{0.2119}
      & \textbf{0.0143} \\
    \bottomrule
  \end{tabular}

  \vspace{0.6em}

  \textbf{(b) Top-$k$ skill selection}\par\vspace{2pt}

  \begin{tabular}{@{}llcccc@{}}
    \toprule
    \textbf{Method}
      & \textbf{Metric}
      & \textbf{Top-1}
      & \textbf{Top-3}
      & \textbf{Top-5}
      & \textbf{Top-10} \\
    \midrule
    Baseline
      & Perf.
      & 79.60\%
      & 79.60\%
      & 79.60\%
      & 79.20\% \\
    & Percentile
      & 51.82
      & 51.82
      & 51.82
      & 43.64 \\
    Full-text
      & Perf.
      & 78.00\%
      & 79.56\%
      & 79.47\%
      & 79.73\% \\
    & Percentile
      & 24.62
      & 50.88
      & 49.02
      & 54.62 \\
    MCRI-Eval
      & Perf.
      & \textbf{80.67\%}
      & 79.56\%
      & \textbf{80.00\%}
      & \textbf{80.13\%} \\
    & Percentile
      & \textbf{74.62}
      & 50.88
      & \textbf{60.23}
      & \textbf{63.11} \\
    \bottomrule
  \end{tabular}

\end{table}

\begin{table}[t]
\caption{Downstream evaluation results on BigCodeBench.
  Top-$k$ results report downstream performance (Perf.) and the
  corresponding percentile score within the candidate set.}
\label{tab:bigcodebench-results}
  \centering
  \footnotesize
  \setlength{\tabcolsep}{3pt}

  \textbf{(a) Ranking agreement}\par\vspace{2pt}

  \begin{tabular}{@{}lccc@{}}
    \toprule
    \textbf{Method}
      & \textbf{C-index}
      & \textbf{Spearman $\rho$}
      & \textbf{$p$-value} \\
    \midrule
    Baseline
      & 0.5686
      & 0.1824
      & 0.0988 \\
    Full-text
      & 0.5648
      & 0.1891
      & 0.0869 \\
    MCRI-Eval
      & \textbf{0.6237}
      & \textbf{0.3355}
      & \textbf{0.0019} \\
    \bottomrule
  \end{tabular}

  \vspace{0.6em}

  \textbf{(b) Top-$k$ skill selection}\par\vspace{2pt}

  \begin{tabular}{@{}llcccc@{}}
    \toprule
    \textbf{Method}
      & \textbf{Metric}
      & \textbf{Top-1}
      & \textbf{Top-3}
      & \textbf{Top-5}
      & \textbf{Top-10} \\
    \midrule
    Baseline
      & Perf.
      & 53.33\%
      & 55.78\%
      & 55.33\%
      & 55.33\% \\
    & Percentile
      & 35.98
      & 75.81
      & 71.34
      & 71.34 \\
    Full-text
      & Perf.
      & 54.67\%
      & 54.22\%
      & 55.20\%
      & 54.60\% \\
    & Percentile
      & 60.37
      & 50.61
      & 69.15
      & 58.90 \\
    MCRI-Eval
      & Perf.
      & \textbf{56.00\%}
      & \textbf{56.00\%}
      & \textbf{56.40\%}
      & \textbf{55.67\%} \\
    & Percentile
      & \textbf{78.05}
      & \textbf{78.05}
      & \textbf{82.07}
      & \textbf{74.70} \\
    \bottomrule
  \end{tabular}

\end{table}

\begin{table}[t]
\caption{Downstream evaluation results on Mind2Web.
  Top-$k$ results report Step Success Rate (Perf.) and the
  corresponding percentile score within the candidate set.}
\label{tab:mind2web-results}
  \centering
  \footnotesize
  \setlength{\tabcolsep}{3pt}

  \textbf{(a) Ranking agreement}\par\vspace{2pt}

  \begin{tabular}{@{}lccc@{}}
    \toprule
    \textbf{Method}
      & \textbf{C-index}
      & \textbf{Spearman $\rho$}
      & \textbf{$p$-value} \\
    \midrule
    Baseline
      & 0.5781
      & 0.1931
      & 0.09687 \\
    Full-text
      & 0.6065
      & 0.2754
      & 0.01678 \\
    MCRI-Eval
      & \textbf{0.6418}
      & \textbf{0.3611}
      & \textbf{0.001458} \\
    \bottomrule
  \end{tabular}

  \vspace{0.6em}

  \textbf{(b) Top-$k$ skill selection}\par\vspace{2pt}

  \begin{tabular}{@{}llcccc@{}}
    \toprule
    \textbf{Method}
      & \textbf{Metric}
      & \textbf{Top-1}
      & \textbf{Top-3}
      & \textbf{Top-5}
      & \textbf{Top-10} \\
    \midrule
    Baseline
      & Perf.
      & 54.88\%
      & 54.88\%
      & 54.90\%
      & 55.14\% \\
    & Percentile
      & 49.55
      & 49.55
      & 50.14
      & 57.55 \\
    Full-text
      & Perf.
      & 56.23\%
      & 56.14\%
      & 55.30\%
      & 55.51\% \\
    & Percentile
      & 79.73
      & 77.93
      & 62.43
      & 66.89 \\
    MCRI-Eval
      & Perf.
      & \textbf{57.39\%}
      & \textbf{56.71\%}
      & \textbf{56.14\%}
      & \textbf{55.54\%} \\
    & Percentile
      & \textbf{99.32}
      & \textbf{92.79}
      & \textbf{78.11}
      & \textbf{67.43} \\
    \bottomrule
  \end{tabular}

\end{table}

\paragraph{Overall results.}
MCRI-Eval achieves the highest C-index and Spearman correlation in all three domains, indicating that the structured evaluation more accurately recovers the ordering of downstream performance than either holistic baseline. The top-$k$ results show the practical consequence of this ranking improvement. On BFCL-Fundamental, the top-1 raw scores are close (80.67\% vs 79.60\%), but MCRI-Eval selects a skill with a normalized rank score of 74.62, compared with 51.82 for the strongest baseline---a gap of 22.80 points. On BigCodeBench, the corresponding scores are 78.05 and 60.37, a gap of 17.68. On Mind2Web, its top-1 selection reaches a Step Success Rate of 57.39\% and a normalized rank score of 99.32, compared with 56.23\% and 79.73 for the strongest baseline---a gap of 19.59.

Thus, the downstream evidence should be interpreted as improved prioritization
within concentrated candidate pools, rather than as a large increase in
absolute task accuracy. MCRI-Eval cannot replace execution-based evaluation,
but it can narrow the search space so that such evaluation focuses on more
promising candidates.

\subsubsection{Stability Analysis}
\label{subsubsec:stability-analysis}

\begin{table}[t]
\caption{Standard deviations of the four MCRI dimension scores.}
\label{tab:mcri-dimension-stability}
  \centering
  \footnotesize
  \setlength{\tabcolsep}{3pt}

  \begin{tabular}{@{}lccccc@{}}
    \toprule
    \textbf{Benchmark}
      & \textbf{$M$ SD}
      & \textbf{$C$ SD}
      & \textbf{$R$ SD}
      & \textbf{$I$ SD}
      & \textbf{\shortstack{Mean\\SD}} \\
    \midrule
    Mind2Web
      & 0.113
      & 0.119
      & 0.118
      & 0.103
      & 0.113 \\
    BigCodeBench
      & 0.139
      & 0.147
      & 0.163
      & 0.141
      & 0.148 \\
    BFCL-Fundamental
      & 0.144
      & 0.107
      & 0.119
      & 0.158
      & 0.132 \\
    \bottomrule
  \end{tabular}

\end{table}

\begin{table}[t]
\caption{Score stability of the three skill evaluation methods.
  The final two columns report the relative reduction in standard
  deviation achieved by MCRI-Eval.}
\label{tab:evaluation-method-stability}
  \centering
  \footnotesize
  \setlength{\tabcolsep}{2.5pt}

  \begin{tabular}{@{}lccccc@{}}
    \toprule
    \textbf{Benchmark}
      & \textbf{Baseline}
      & \textbf{Full-text}
      & \textbf{MCRI}
      & ${\Delta}_{\mathrm{Base}}$
      & ${\Delta}_{\mathrm{Full}}$ \\
    \midrule
    Mind2Web
      & 0.119
      & 0.197
      & \textbf{0.085}
      & \textbf{28.5\%}
      & \textbf{56.7\%} \\
    BigCodeBench
      & 0.476
      & 0.396
      & \textbf{0.061}
      & \textbf{87.1\%}
      & \textbf{84.5\%} \\
    BFCL-Fundamental
      & 0.095
      & 0.280
      & \textbf{0.078}
      & \textbf{18.4\%}
      & \textbf{72.3\%} \\
    Cross-bench.\ avg.
      & 0.230
      & 0.291
      & \textbf{0.075}
      & \textbf{67.5\%}
      & \textbf{74.3\%} \\
    \bottomrule
  \end{tabular}

\end{table}

\paragraph{Stability results.}
The standard deviations of the four MCRI dimension scores range from 0.103 to 0.163, indicating that the dimension-level evaluations are stable across repeated assessments. MCRI-Eval achieves the lowest overall standard deviation in all three domains and is therefore more stable than both the Baseline and the Full-text Holistic Baseline.

On average across the three benchmarks, MCRI-Eval reduces the standard deviation by 67.5\% relative to the Baseline and by 74.3\% relative to the Full-text Holistic Baseline. These results suggest that decomposing skill quality into explicitly defined dimensions substantially reduces the variance associated with unstructured holistic evaluation.

\section{Discussion and Conclusion}

We propose MCRI as an analytical framework for skills and evaluate its utility through community indicators and downstream task performance.

\paragraph{Why there is no ``ground truth'' of quality.}
Skill quality is 
definition-dependent: the same skill may be valued differently across tasks, models, and user preferences. Consequently, no task-independent quality value is available for MCRI to fit directly. We instead examine whether its scores exhibit expected associations with external signals.

\paragraph{Positioning the two proxy indicators.}
The two indicators capture different aspects. Community indicators measure popularity, whereas downstream performance is closer to task-specific utility but remains sensitive to task--skill alignment, sample size, and evaluation methodology.

MCRI therefore does not fully define skill quality; it provides a mechanism-driven, operationalizable, and extensible analytical lens for pre-execution selection.

%% file: supplement_body.tex
\section{Additional Related Work}
\label{sec:related}

Research on agent skills is still at an early stage. As of July 28, 2026, the
closely related studies identified in our search were available as arXiv
preprints, and we found no evidence in the supplied materials that they had
been formally published or accepted.

Existing work can be organized into four lines. First, benchmark-oriented
studies evaluate whether skills improve agent performance through actual task
execution. Representative examples include skillsBench
\cite{li2026skillsbench}, evaluations of skill use in realistic settings
\cite{liu2026wild}, a scalable evaluation framework
\cite{shaposhnikov2026framework}, and OpenskillEval
\cite{ying2026openskilleval}. Second, research on skill generation and
improvement extracts transferable experience from agent trajectories; an
example is Trace2skill \cite{ni2026trace2skill}. Third, survey work organizes
the emerging paradigms for skill evaluation and evolution
\cite{ding2026evaluation}. Fourth, security-oriented studies audit open skills
for malicious behavior or multidimensional risk, as illustrated by skillSieve
\cite{hou2026skillsieve} and skillVetBench
\cite{hossain2026skillvetbench}.

Together, these studies cover skill construction, execution-based evaluation,
and security auditing. They leave three issues comparatively underexplored:
the mechanisms through which a skill affects an agent, theoretically grounded
criteria for evaluating skill quality, and a general pre-execution evaluator
that can support skill selection without running every candidate.

\section{Dataset Construction and Filtering}
\label{sec:dataset}

The ecosystem-level data were collected from the OpenClaw skill Hub API
(\url{https://clawhub.ai/api/v1/skills}) on May 5, 2026. The raw index contained
63,820 records, of which 63,812 remained after removing invalid or empty
skills. Each record provides basic metadata (\texttt{slug}, \texttt{name},
\texttt{summary}, \texttt{createdAt}, \texttt{updatedAt}, and
\texttt{latestVersion}) and observable platform signals (\texttt{stars}, 
\texttt{downloads}, \texttt{installsAllTime}, 
\texttt{installsCurrent},
\texttt{comments}, and \texttt{versions}).

Raw download and installation counts are jointly affected by time since
publication and platform exposure. We therefore use downloads per day,
\[
  \text{downloads per day}
  =
  \frac{\text{downloads}}
       {\text{reference date}-\texttt{createdAt}},
\]
in the ecosystem-level correlation analysis. Because the platform signals are
strongly long-tailed, the analysis focuses on the top 3\%, 1\%, and 0.5\%
subsets, corresponding to $N=1{,}823$, 606, and 305. The experimental resource
package described in Appendix~\ref{sec:provenance} contains the filtered $N=1{,}823$ subset and the exact $N=606$ and
$N=305$ threshold definitions used in the analysis.

\section{Supplementary Experiments}
\label{sec:supp_experiments}

We report three additional evaluations of MCRI-Eval as a pre-execution skill
quality estimator. The first compares MCRI-Eval with multi-stage heuristic
scorers that have comparable or larger reasoning budgets. The second measures
downstream performance without injecting any skill. The third compares
MCRI-Eval with selectors based only on community signals, text length, or file
count. These evaluations reuse the BFCL-Fundamental, BigCodeBench, and
Mind2Web settings from the main experiments; no new task set or evaluation
metric is introduced here.

\subsection{Common Evaluation Protocol}
\label{sec:common_protocol}

Each method assigns a selector score to every skill. The corresponding
outcome is the downstream performance obtained after injecting that skill:
\texttt{pass\_rate} for BFCL-Fundamental and BigCodeBench, and Step Success
Rate (Step SR) for Mind2Web. When a method produces multiple scores and
confidence values, its selector is the product of the two means,
\[
  \text{selector}
  =
  \operatorname{mean}(\text{score})
  \times
  \operatorname{mean}(\text{confidence}).
\]

We report the concordance index (C-index), Spearman's rank correlation
$\rho$ and its $p$-value, and Top-$k$ selection. A C-index of 0.5 corresponds
to random pairwise ordering. For Top-$k$, we report both the mean downstream
outcome among the selected skills and a normalized rank score. The rank score
locates that mean outcome in the empirical distribution of per-skill outcomes,
with 0 denoting the worst position and 100 the best. Ties at the selector
cutoff are handled by the expected Top-$k$ outcome.

\subsection{Multi-Stage Heuristic Scorers}
\label{sec:heuristics}

The heuristic comparison controls for the possibility that MCRI-Eval benefits
only from a larger reasoning budget. In stage one, the model reads the complete
skill text and the benchmark description. Three independent runs each produce
an initial quality score, an initial confidence, and an unrestricted
evaluation. In stage two, the model receives the skill, the task, and a
stage-one evaluation, then returns a revised score and confidence. The final
confidence for a run is the product of the stage-one and stage-two
confidences.

The two variants share the same three stage-one evaluations. In
\texttt{heuristic\_a}, the evaluations are summarized into one intermediate
assessment; their mean initial score and confidence are attached, and three
stage-two runs are performed on that summary. In \texttt{heuristic\_b}, the
$i$-th stage-one evaluation is passed directly to the $i$-th stage-two run.
Thus, \texttt{heuristic\_a} uses an additional summarization step, whereas
\texttt{heuristic\_b} preserves one-to-one pairing. Each selector is computed
from the three final scores and final confidence products.

\begin{table}[!t]
\caption{Ranking agreement of the heuristic comparison. BFCL, BCB, and M2W
denote BFCL-Fundamental, BigCodeBench, and Mind2Web, respectively. MCRI-Eval
attains the highest C-index and Spearman correlation on all three benchmarks.}
\label{tab:heuristic-ranking}
\centering
\footnotesize
\setlength{\tabcolsep}{2.0pt}
\begin{tabular}{llrrr}
\toprule
Benchmark & Method & C-ind. & $\rho$ & $p$ \\
\midrule
BFCL
  & Baseline & 0.5101 & 0.0248 & 0.7765 \\
  & FullTextBaseline & 0.5522 & 0.1496 & 0.0856 \\
  & \texttt{heuristic\_a} & 0.5682 & 0.1846 & 0.0334 \\
  & \texttt{heuristic\_b} & 0.5699 & 0.1849 & 0.0331 \\
  & \textbf{MCRI-Eval} & \textbf{0.5777} & \textbf{0.2119} & \textbf{0.0143} \\
\midrule
BCB
  & Baseline & 0.5686 & 0.1824 & 0.0988 \\
  & FullTextBaseline & 0.5648 & 0.1891 & 0.0869 \\
  & \texttt{heuristic\_a} & 0.5678 & 0.1876 & 0.0894 \\
  & \texttt{heuristic\_b} & 0.5670 & 0.1848 & 0.0944 \\
  & \textbf{MCRI-Eval} & \textbf{0.6237} & \textbf{0.3355} & \textbf{0.0019} \\
\midrule
M2W
  & Baseline & 0.5781 & 0.1931 & 0.0969 \\
  & FullTextBaseline & 0.6065 & 0.2754 & 0.0168 \\
  & \texttt{heuristic\_a} & 0.5602 & 0.1689 & 0.1473 \\
  & \texttt{heuristic\_b} & 0.5642 & 0.1812 & 0.1198 \\
  & \textbf{MCRI-Eval} & \textbf{0.6418} & \textbf{0.3611} & \textbf{0.001458} \\
\bottomrule
\end{tabular}

\end{table}

\begin{table*}[!t]
\caption{Top-$k$ results for the heuristic comparison. Each cell reports
downstream performance (\%) / normalized rank score. Bold marks the highest
downstream performance for each benchmark and value of $k$.}
\label{tab:heuristic-topk}
\centering
\footnotesize
\setlength{\tabcolsep}{2.6pt}
\begin{tabular}{llrrrrr}
\toprule
Benchmark & Method & Top-1 & Top-3 & Top-5 & Top-10 & Top-20 \\
\midrule
BFCL
  & Baseline & 79.60 / 51.82 & 79.60 / 51.82 & 79.60 / 51.82 & 79.20 / 43.64 & 78.80 / 35.91 \\
  & FullTextBaseline & 78.00 / 24.62 & 79.56 / 50.88 & 79.47 / 49.02 & 79.73 / 54.62 & 79.37 / 46.91 \\
  & \texttt{heuristic\_a} & 78.67 / 33.33 & 79.33 / 46.21 & 79.73 / 54.62 & 79.20 / 43.64 & 79.33 / 46.21 \\
  & \texttt{heuristic\_b} & 78.67 / 33.33 & \textbf{80.00 / 60.23} & \textbf{80.00 / 60.23} & 79.13 / 42.35 & \textbf{79.60 / 51.82} \\
  & MCRI-Eval & \textbf{80.67 / 74.62} & 79.56 / 50.88 & \textbf{80.00 / 60.23} & \textbf{80.13 / 63.11} & 79.28 / 45.25 \\
\midrule
BigCodeBench
  & Baseline & 53.33 / 35.98 & 55.78 / 75.81 & 55.33 / 71.34 & 55.33 / 71.34 & 55.00 / 65.85 \\
  & FullTextBaseline & 54.67 / 60.37 & 54.22 / 50.61 & 55.20 / 69.15 & 54.60 / 58.90 & 55.00 / 65.85 \\
  & \texttt{heuristic\_a} & 54.67 / 60.37 & 54.00 / 45.73 & 54.27 / 51.59 & 55.13 / 68.05 & \textbf{55.13 / 68.05} \\
  & \texttt{heuristic\_b} & 54.67 / 60.37 & 54.00 / 45.73 & 54.93 / 64.76 & 55.27 / 70.24 & 54.93 / 64.76 \\
  & MCRI-Eval & \textbf{56.00 / 78.05} & \textbf{56.00 / 78.05} & \textbf{56.40 / 82.07} & \textbf{55.67 / 74.70} & 55.07 / 66.95 \\
\midrule
Mind2Web
  & Baseline & 54.88 / 49.55 & 54.88 / 49.55 & 54.90 / 50.14 & 55.14 / 57.55 & 55.43 / 65.54 \\
  & FullTextBaseline & 56.23 / 79.73 & 56.14 / 77.93 & 55.30 / 62.43 & 55.51 / 66.89 & 55.38 / 64.46 \\
  & \texttt{heuristic\_a} & 56.23 / 79.73 & 56.04 / 76.13 & 55.65 / 69.59 & 55.36 / 64.19 & 55.35 / 63.75 \\
  & \texttt{heuristic\_b} & 56.23 / 79.73 & 55.85 / 72.75 & 55.65 / 69.59 & 55.36 / 64.19 & 55.39 / 64.73 \\
  & MCRI-Eval & \textbf{57.39 / 99.32} & \textbf{56.71 / 92.79} & \textbf{56.14 / 78.11} & \textbf{55.54 / 67.43} & \textbf{55.52 / 67.16} \\
\bottomrule
\end{tabular}

\end{table*}

MCRI-Eval yields the highest C-index and Spearman correlation in every
benchmark and is the only method with $p<0.05$ in all three. The heuristic
variants do not consistently improve over FullTextBaseline despite using
additional reasoning stages. The comparison therefore does not support an
explanation based solely on the number of inference steps; the structured MCRI
decomposition provides a more stable ranking signal in these settings.

\subsection{No-skill Reference}
\label{sec:noskill}

The No-skill condition measures downstream performance when the prompt
contains only the task or function definition and no skill text. The fixed
50-task set of each benchmark is repeated three times. BigCodeBench and
Mind2Web use \texttt{deepseek-v4-flash} with reasoning disabled, while
BFCL-Fundamental uses \texttt{gpt-5.4-nano}, matching the corresponding
resource files. Because the No-skill condition assigns no selector, it has no
ranking-agreement statistics. Its downstream level is instead located in the
distribution of skill-injected outcomes.

\begin{table*}[!t]
\caption{No-skill downstream performance over three repeated runs. The final
column is the percentile of the mean in the corresponding distribution of
skill-injected outcomes.}
\label{tab:noskill}
\centering
\footnotesize
\setlength{\tabcolsep}{3pt}
\begin{tabular}{@{}lcc@{}}
\toprule
Benchmark & R1 / R2 / R3 & Mean / Rank \\
\midrule
BFCL-Fundamental & 0.800 / 0.800 / 0.800 & 80.00\% / 54.00 \\
BigCodeBench & 0.560 / 0.540 / 0.500 & 53.33\% / 35.98 \\
Mind2Web & 0.519 / 0.533 / 0.513 & 52.17\% / 6.05 \\
\bottomrule
\end{tabular}

\end{table*}

The usefulness of injecting a skill depends on the benchmark. On Mind2Web,
the No-skill result lies near the bottom of the skill-injected distribution,
whereas the skill selected by MCRI-Eval reaches the 99.32 rank score. On
BigCodeBench, the No-skill mean is lower than the MCRI-Eval Top-1 outcome.
On BFCL-Fundamental, the No-skill mean is slightly above the average
skill-injected level reported in the resource summary, yet the MCRI-Eval
Top-1 outcome remains higher. These results motivate pre-execution selection:
injecting an arbitrary skill is not uniformly beneficial, while selecting a
task-suitable skill can improve the outcome relative to using no skill.

\subsection{Non-LLM Selectors}
\label{sec:nonllm}

The main Baseline, FullTextBaseline, and MCRI-Eval methods all invoke an LLM.
To test whether MCRI-Eval merely recovers inexpensive surface signals, we add
five non-LLM selectors belonging to three categories. Community-based
selection uses raw \texttt{stars}, \texttt{downloads}, or
\texttt{installsAllTime} from the supplied per-skill metadata snapshots.
Missing values are set to zero; all benchmark skills have matching snapshots
(133/133 for BFCL-Fundamental, 83/83 for BigCodeBench, and 75/75 for
Mind2Web). Text-length selection uses the character count of the exact
\texttt{<slug>\_text.txt} content injected into the model. File-count
selection recursively counts files in the packaged skill directory as a
coarse proxy for resource richness.

These selectors use the same downstream outcomes and evaluation procedure as
the LLM-based methods. Community signals are intentionally not normalized by
skill age in this experiment: the question here is whether the observable
signal itself can select skills with stronger downstream outcomes, rather
than whether it is an age-adjusted measure of ecosystem popularity.

\begin{table*}[!t]
\caption{Non-LLM selector comparison on BFCL-Fundamental ($N=133$).
Top-$k$ cells report pass rate (\%) / normalized rank score. MCRI-Eval has
the strongest positive global ranking agreement; the file-count Top-1 value
is an isolated high-outcome selection despite its negative overall
correlation.}
\label{tab:nonllm-bfcl}
\centering
\footnotesize
\setlength{\tabcolsep}{1.6pt}
\begin{tabular}{lrrrrrrr}
\toprule
Method & C-index & $\rho$ & $p$-value & Top-1 & Top-3 & Top-5 & Top-10 \\
\midrule
Baseline (LLM) & 0.5101 & 0.0248 & 0.7765 & 79.60 / 51.82 & 79.60 / 51.82 & 79.60 / 51.82 & 79.20 / 43.64 \\
FullTextBaseline (LLM) & 0.5522 & 0.1496 & 0.0856 & 78.00 / 24.62 & 79.56 / 50.88 & 79.47 / 49.02 & 79.73 / 54.62 \\
\textbf{MCRI-Eval (LLM)} & \textbf{0.5777} & \textbf{0.2119} & \textbf{0.0143} & 80.67 / 74.62 & \textbf{79.56 / 50.88} & \textbf{80.00 / 60.23} & \textbf{80.13 / 63.11} \\
Stars & 0.5107 & 0.0315 & 0.7186 & 76.67 / 15.15 & 76.89 / 16.41 & 78.93 / 38.48 & 78.87 / 37.20 \\
Downloads & 0.5161 & 0.0467 & 0.5938 & 76.67 / 15.15 & 77.33 / 18.94 & 78.93 / 38.48 & 78.93 / 38.48 \\
All-time installs & 0.4430 & -0.1557 & 0.0736 & 76.00 / 10.61 & 77.33 / 18.94 & 78.93 / 38.48 & 78.40 / 29.85 \\
skill text length & 0.4987 & 0.0058 & 0.9468 & 74.67 / 4.55 & 77.33 / 18.94 & 77.87 / 23.48 & 78.07 / 25.49 \\
skill file count & 0.4253 & -0.2237 & 0.0096 & \textbf{82.00 / 93.56} & 78.89 / 37.63 & 78.27 / 28.11 & 78.27 / 28.11 \\
\bottomrule
\end{tabular}

\end{table*}

\begin{table*}[!t]
\caption{Non-LLM selector comparison on BigCodeBench ($N=83$). Top-$k$
cells report pass rate (\%) / normalized rank score. Text length is
predictive in this benchmark, but MCRI-Eval retains the strongest global
ranking agreement.}
\label{tab:nonllm-bcb}
\centering
\footnotesize
\setlength{\tabcolsep}{1.6pt}
\begin{tabular}{lrrrrrrr}
\toprule
Method & C-index & $\rho$ & $p$-value & Top-1 & Top-3 & Top-5 & Top-10 \\
\midrule
Baseline (LLM) & 0.5686 & 0.1824 & 0.0988 & 53.33 / 35.98 & 55.78 / 75.81 & 55.33 / 71.34 & 55.33 / 71.34 \\
FullTextBaseline (LLM) & 0.5648 & 0.1891 & 0.0869 & 54.67 / 60.37 & 54.22 / 50.61 & 55.20 / 69.15 & 54.60 / 58.90 \\
\textbf{MCRI-Eval (LLM)} & \textbf{0.6237} & \textbf{0.3355} & \textbf{0.0019} & 56.00 / 78.05 & \textbf{56.00 / 78.05} & \textbf{56.40 / 82.07} & \textbf{55.67 / 74.70} \\
Stars & 0.5093 & 0.0321 & 0.7733 & 54.67 / 60.37 & 54.00 / 45.73 & 53.47 / 37.93 & 55.03 / 66.40 \\
Downloads & 0.4762 & -0.0646 & 0.5620 & 53.33 / 35.98 & 54.00 / 45.73 & 53.60 / 39.88 & 53.93 / 44.76 \\
All-time installs & 0.5238 & 0.0679 & 0.5418 & 53.33 / 35.98 & 52.89 / 31.10 & 53.60 / 39.88 & 54.80 / 62.56 \\
skill text length & 0.6012 & 0.2658 & 0.0152 & \textbf{58.00 / 90.85} & \textbf{56.00 / 78.05} & 55.60 / 74.02 & 55.27 / 70.24 \\
skill file count & 0.5462 & 0.1457 & 0.1888 & 54.00 / 45.73 & 53.56 / 39.23 & 53.47 / 37.93 & 53.80 / 42.80 \\
\bottomrule
\end{tabular}

\end{table*}

\begin{table*}[!t]
\caption{Non-LLM selector comparison on Mind2Web ($N=75$). Top-$k$ cells
report Step SR (\%) / normalized rank score. MCRI-Eval provides the strongest
global ranking agreement and the best Top-$k$ outcomes.}
\label{tab:nonllm-m2w}
\centering
\footnotesize
\setlength{\tabcolsep}{1.6pt}
\begin{tabular}{lrrrrrrr}
\toprule
Method & C-index & $\rho$ & $p$-value & Top-1 & Top-3 & Top-5 & Top-10 \\
\midrule
Baseline (LLM) & 0.5781 & 0.1931 & 0.0969 & 54.88 / 49.55 & 54.88 / 49.55 & 54.90 / 50.14 & 55.14 / 57.55 \\
FullTextBaseline (LLM) & 0.6065 & 0.2754 & 0.0168 & 56.23 / 79.73 & 56.14 / 77.93 & 55.30 / 62.43 & 55.51 / 66.89 \\
\textbf{MCRI-Eval (LLM)} & \textbf{0.6418} & \textbf{0.3611} & \textbf{0.001458} & \textbf{57.39 / 99.32} & \textbf{56.71 / 92.79} & \textbf{56.14 / 78.11} & \textbf{55.54 / 67.43} \\
Stars & 0.4508 & -0.1450 & 0.2144 & 56.52 / 87.84 & 53.72 / 23.65 & 54.55 / 39.59 & 54.41 / 35.61 \\
Downloads & 0.4916 & -0.0123 & 0.9169 & 56.52 / 87.84 & 54.98 / 52.48 & 54.20 / 30.41 & 54.26 / 31.89 \\
All-time installs & 0.5206 & 0.0578 & 0.6226 & 56.52 / 87.84 & 53.72 / 23.65 & 54.20 / 30.41 & 54.61 / 41.35 \\
skill text length & 0.4869 & -0.0400 & 0.7335 & 54.20 / 30.41 & 54.01 / 27.70 & 53.80 / 24.73 & 54.55 / 39.59 \\
skill file count & 0.5071 & 0.0139 & 0.9058 & 55.07 / 55.41 & 54.59 / 40.77 & 54.26 / 31.89 & 54.87 / 49.26 \\
\bottomrule
\end{tabular}

\end{table*}

MCRI-Eval has the highest C-index and Spearman correlation among all compared
selectors on every benchmark, and it is the only method whose positive
association is significant in all three. Community popularity is a weak
predictor of downstream task performance: its correlations fall between
$-0.1557$ and $0.0679$. This does not conflict with a positive relationship
between MCRI quality and ecosystem popularity, because popularity and
task-specific post-injection performance are distinct criteria.

Text length is a noteworthy exception on BigCodeBench
($\rho=0.2658$, $p=0.0152$). A plausible interpretation is that longer
code-oriented skills can contain more reusable examples and library-specific
information, partially overlapping the Resources dimension. However, the same
signal vanishes on BFCL-Fundamental ($\rho=0.0058$) and becomes slightly
negative on Mind2Web ($\rho=-0.0400$), so it does not support the general rule
that longer skills are better. File count is significantly negative on
BFCL-Fundamental ($\rho=-0.2237$, $p=0.0096$). Because the evaluation injects
skill documents as text without executing bundled scripts, a package that
delegates important behavior to external files may provide less directly
usable information in this setting. The high BFCL Top-1 value for file count
is therefore a single selected outcome and does not reverse the negative
global ordering.

\subsection{Summary of the Supplementary Evidence}
\label{sec:supp_summary}

The three comparisons support a bounded conclusion. Increasing the number of
reasoning stages does not reproduce MCRI-Eval's ranking consistency; omitting
skills shows that arbitrary injection is not uniformly helpful; and readily
computed surface signals are either weak, task-specific, or unstable. Across
all three benchmarks, MCRI-Eval is the only selector with a positive and
statistically significant Spearman association in every setting.

\section{Complete MCRI-Eval Definition and Scoring Rubric}
\label{sec:rubric}

MCRI-Eval has two stages. Stage A independently scores the Metadata,
Constraints, Resources, and Instructions dimensions on a 1--5 scale. Stage B
combines the four dimension scores with the benchmark description and asks an
LLM for a final quality score and confidence. The selector is the product of
the mean final score and mean final confidence across repeated runs.

\subsection{Dimension Definitions}

\begin{table}[H]
\caption{The four MCRI dimensions and their functional roles.}
\label{tab:mcri-definitions}
\centering
\footnotesize
\begin{tabularx}{\textwidth}{@{}c l l >{\raggedright\arraybackslash}X@{}}
\toprule
Dim. & Name & Functional layer & Definition \\
\midrule
M & Metadata & Identification &
Task definition, functional scope, applicable scenarios, triggering
conditions, and declarative input/output specifications that tell the host
agent what the skill does and when it should be used. \\
C & Constraints & Constraint &
Rules that explicitly restrict the model's admissible behavior, including
prohibitions, preconditions, invariants, handling policies, and boundary
conditions. \\
R & Resources & Information &
Task-specific information beyond parametric knowledge, including templates,
scripts, example data, reference documents, concrete API details, and
previously observed error traces. \\
I & Instructions & Interface &
Executable guidance for invoking internal scripts, external tools, or
sub-skills and interacting with the environment, including call forms,
orchestration order, and subtask decomposition. \\
\bottomrule
\end{tabularx}

\end{table}

\FloatBarrier
\subsection{Stage A: Dimension Scoring}

The experiments used a Chinese system prompt. Listing~\ref{lst:stage-a}
provides a faithful English rendering of the readable prompt distributed in
the resource package; the field names and output schema are unchanged.

\Needspace{12\baselineskip}
\begin{lstlisting}[caption={English rendering of the Stage-A system prompt.},label={lst:stage-a}]
You are a strict, evidence-oriented evaluator of AI skill quality. Score the skill using the MCRI framework based only on the supplied skill name, summary, and complete text. Do not use any benchmark, community popularity, downstream performance, or external impression of the skill.

[Academic definitions of MCRI]
M (Metadata; identification layer): describes the task definition, functional scope, applicable scenarios, triggering conditions, and input/output specifications, allowing the host agent to know what the skill does and when it should be used.
C (Constraints; constraint layer): explicitly restricts the model's admissible behavior through prohibitions, preconditions, invariants, handling policies, and boundary conditions.
R (Resources; information layer): injects information beyond the model's parametric knowledge that can be directly used for the task, including templates, scripts, example data, reference documents, concrete API details, and error traces.
I (Instructions; interface layer): specifies how to invoke internal scripts, external tools, and sub-skills and how to interact with the environment, including invocation forms, orchestration, and subtask decomposition.

[Construct boundaries]
M evaluates the declared task and when the skill should be used; I evaluates how execution is invoked and orchestrated. A task description alone does not justify a high I score.
C counts only rules, conditions, and invariants that exclude erroneous or off-target behavior. General advice, factual descriptions, and positive workflow steps do not count as C.
R counts only task-specific, reusable information that is actually supplied. Merely naming a tool, giving an external link, describing a product capability, or stating common knowledge does not provide high information gain.
I counts executable interfaces and orchestration. Merely saying "use a tool" without a call contract, parameters, ordering, or result handling does not justify a high score.
Do not reward length, repetition, or polished formatting.

[Unified scale for each sub-item: integer from 0 to 100]
0: completely absent, contradictory, or only an unverifiable assertion.
25: merely mentioned and highly ambiguous.
50: partially specified but important details still require inference.
75: clear, specific, and largely operational, with only limited substantive omissions.
100: comprehensive, precise, directly executable or verifiable, and without important omissions.
Any integer in [0,100] may be used; do not default to multiples of ten. Score missing sub-items honestly. Do not penalize brevity or reward length. Do not infer the contents of links or undisclosed files. Evaluate a skill relative to the requirements of the task it claims to address.

[M sub-items]
task_definition: specificity and distinctiveness of the task objective and functional scope.
use_conditions: clarity of positive triggers, applicable scenarios, and when the skill should be selected.
scope_boundaries: clarity of inapplicable cases, capability limits, prerequisites, and scope.
declarative_io: whether required inputs and produced outputs are declared, without double-counting invocation steps.
consistency: consistency and retrievability of the task claim across the name, summary, and body.

[C sub-items]
explicit_rules: explicit MUST/MUST NOT rules or invariants.
decision_coverage: coverage of important task decisions rather than peripheral details.
operability: whether the rules are concrete enough to follow or verify.
failure_boundaries: treatment of failures, exceptions, safety, stopping conditions, and boundary cases.
non_conflict: whether constraints are consistent, necessary, and not excessive.

[R sub-items]
extra_parametric_value: task-specific information that cannot be reliably inferred from general model knowledge.
specificity: precision of APIs, fields, parameters, code, templates, data, or examples.
critical_coverage: coverage of the main information gaps required to complete the task.
reusability: whether the information can be directly reused to reduce search, guessing, or reinvention.
validity: whether conditions of use, versioning, internal consistency, or provenance support reliable reuse.

[I sub-items]
callable_entities: clarity of actual scripts, tools, commands, sub-skills, or interaction actions. If external tools are unnecessary, evaluate whether the actual execution actions are clear.
invocation_contract: clarity of call syntax, parameters, input formats, return values, and prerequisites.
orchestration: executability of ordering, subtask decomposition, branches, and completion conditions.
state_artifacts: clarity of intermediate state, files, artifacts, result transfer, and persistence.
environment_handoff: clarity of environment dependencies, error returns, retries, user confirmation, and handoffs to external components.

Return exactly one JSON object. Do not use Markdown and do not output an overall score or the four 1--5 dimension scores:
{
  "subscores": {
    "M": {"task_definition": 0, "use_conditions": 0, "scope_boundaries": 0, "declarative_io": 0, "consistency": 0},
    "C": {"explicit_rules": 0, "decision_coverage": 0, "operability": 0, "failure_boundaries": 0, "non_conflict": 0},
    "R": {"extra_parametric_value": 0, "specificity": 0, "critical_coverage": 0, "reusability": 0, "validity": 0},
    "I": {"callable_entities": 0, "invocation_contract": 0, "orchestration": 0, "state_artifacts": 0, "environment_handoff": 0}
  },
  "basis": {
    "M": "the most important supporting evidence and omission",
    "C": "the most important supporting evidence and omission",
    "R": "the most important supporting evidence and omission",
    "I": "the most important supporting evidence and omission"
  }
}
\end{lstlisting}

The user message concatenates the skill name, skill summary, and complete text
inside a \texttt{<SKILL\_FULL\_TEXT>} block. Within each dimension, the five
0--100 sub-item scores are aggregated and linearly mapped to 1--5:
\[
  \operatorname{score}_{d}
  =
  1 + 4
  \frac{\sum_i w_{d,i}\operatorname{subscore}_{d,i}}{100}.
\]

\begin{table}[H]
\caption{Stage-A sub-item weights. The weights within each dimension sum to
one.}
\label{tab:mcri-weights}
\centering
\footnotesize
\begin{tabularx}{\textwidth}{@{}c l >{\raggedright\arraybackslash}X@{}}
\toprule
Dim. & Output name & Sub-item weights \\
\midrule
M & \texttt{task\_spec\_clarity} &
\texttt{task\_definition} (0.25), \texttt{use\_conditions} (0.25),
\texttt{scope\_boundaries} (0.20), \texttt{declarative\_io} (0.20),
\texttt{consistency} (0.10) \\
C & \texttt{constraint\_sufficiency} &
\texttt{explicit\_rules} (0.25), \texttt{decision\_coverage} (0.25),
\texttt{operability} (0.20), \texttt{failure\_boundaries} (0.20),
\texttt{non\_conflict} (0.10) \\
R & \texttt{information\_gain} &
\texttt{extra\_parametric\_value} (0.30), \texttt{specificity} (0.25),
\texttt{critical\_coverage} (0.20), \texttt{reusability} (0.15),
\texttt{validity} (0.10) \\
I & \texttt{interface\_completeness} &
\texttt{callable\_entities} (0.20), \texttt{invocation\_contract} (0.25),
\texttt{orchestration} (0.25), \texttt{state\_artifacts} (0.15),
\texttt{environment\_handoff} (0.15) \\
\bottomrule
\end{tabularx}

\end{table}

\subsection{Task-Conditioned Stage-A Variant}

The ecosystem-level community analysis uses the task-independent Stage-A
prompt. The downstream experiments replace its first exclusion sentence with:

\begin{quote}
An MCRI score evaluates a skill for a particular task rather than its
task-independent general quality. Using only the current task, judge whether
the content of each skill dimension is useful for completing that task. Do not
use community popularity, downstream performance, or external impressions.
\end{quote}

The prompt then appends a \texttt{[Current Task]} block. The BigCodeBench block,
for example, states that the benchmark contains function-level Python tasks
involving libraries and operations such as pandas, NumPy, Matplotlib, text and
regular expressions, file processing, basic statistics, and machine learning.
Given imports, a function signature, and a docstring, the model generates a
function that is evaluated by local unit tests. BFCL-Fundamental and Mind2Web
use their corresponding task descriptions from the supplied code. The
construct definitions, sub-items, weights, and output schema remain unchanged.

\FloatBarrier
\subsection{Stage B: Overall Scoring}

Baseline, FullTextBaseline, and MCRI-Eval share the output requirement in
Listing~\ref{lst:stage-b}. The experiments used the Chinese text; the listing
is its faithful English rendering.

\Needspace{12\baselineskip}
\begin{lstlisting}[caption={English rendering of the common Stage-B output requirement.},label={lst:stage-b}]
Output exactly one JSON object in the form {"score": float, "confidence": float}.
score is an overall quality score from 1 to 5, with exactly two decimal places, and it must not be an integer.
confidence is a selection confidence from 0.00 to 1.00, with exactly two decimal places. It is the discount factor applied to score for the current task; it is not the skill quality itself and is not the task completion rate.
Specifically, confidence=1.00 means that the skill fully matches the current task and the quality score is fully reliable, so selection may use score directly. confidence=0.80 means that the skill largely matches but task adaptation or information uncertainty remains, so selection should use score x 0.80. confidence=0.50 means that the skill only partially matches or uncertainty is substantial, so selection should use score x 0.50. confidence=0.20 means that the quality score has little selection value for the current task, so selection should use score x 0.20. Choose the most appropriate value between these anchors.
In the downstream evaluation, the skill document is injected into the language model only as a text prompt. The evaluation does not execute bundled scripts or resource files and does not call external APIs, browsers, network services, or other external tools declared by the skill.
Do not output an explanation, Markdown, or any other text.
\end{lstlisting}

\begin{table}[H]
\caption{Information supplied to each Stage-B method. All other output and
aggregation rules are shared.}
\label{tab:stage-b-methods}
\centering
\footnotesize
\begin{tabularx}{\textwidth}{@{}l >{\raggedright\arraybackslash}X >{\raggedright\arraybackslash}X@{}}
\toprule
Method & Reference information & Additional input \\
\midrule
Baseline &
A statement that the experiment measures the LLM's unaided ability to assess
skill quality. &
skill name, summary, and benchmark task description. \\
FullTextBaseline &
The same reference statement as Baseline. &
Baseline input plus the complete skill text. \\
MCRI-Eval &
The MCRI theoretical description, including information gain, behavioral
constraint, and the four dimension definitions. &
The mean Stage-A values
$M=x.xx$, $C=x.xx$, $R=x.xx$, and $I=x.xx$, accompanied by the instruction
that dimension importance can vary by benchmark and that the values are
advisory rather than a fixed aggregation formula. \\
\bottomrule
\end{tabularx}

\end{table}

For each method, three final runs are aggregated as
\[
  \text{selector}
  =
  \operatorname{mean}(\text{final score})
  \times
  \operatorname{mean}(\text{final confidence}).
\]

\section{Dataset Concentration and Distributions}
\label{sec:distributions}

\subsection{Concentration of Downstream Outcomes}

The per-skill downstream outcomes are concentrated in narrow intervals:

\begin{itemize}
  \item \textbf{BFCL-Fundamental} ($N=133$): pass rate ranges from 73.33\% to
  83.33\%, with mean 79.15\% and standard deviation 2.17 percentage points. A
  five-percentage-point sliding window contains 79.7\% of the skills.
  \item \textbf{BigCodeBench} ($N=83$): pass rate ranges from 47.33\% to
  60.00\%, with mean 54.13\% and standard deviation 2.51 percentage points.
  The densest five-point window contains 71.1\% of the skills.
  \item \textbf{Mind2Web} ($N=75$): Step SR ranges from 50.14\% to 57.39\%,
  with mean 54.79\% and standard deviation 1.49 percentage points. The densest
  five-point window contains 93.3\% of the skills.
\end{itemize}

This concentration motivates reporting both raw downstream performance and
the normalized rank score. When most skills occupy a band only a few
percentage points wide, a difference of 0.5 percentage points can correspond
to many positions in the per-skill ordering; raw percentages alone can
therefore understate differences in selection quality.

\subsection{Skill Text Length and File Count}

The injected skill texts are right-skewed in all three benchmark subsets.
BFCL-Fundamental has mean length 19,023 characters, median 15,402, 90th
percentile 37,391, and maximum 131,424. BigCodeBench has mean 19,429,
median 13,732, 90th percentile 45,363, and maximum 97,697. Mind2Web has
mean 19,428, median 14,028, 90th percentile 41,193, and maximum 75,299.
Thus, the medians are close to 14,000--15,000 characters, while a small number
of skills are substantially longer.

Packaged file counts are also right-skewed but much smaller. The mean,
median, 90th percentile, and maximum are respectively 2.72, 1, 7, and 16 for
BFCL-Fundamental; 3.46, 3, 7, and 14 for BigCodeBench; and 4.05, 3, 9, and
25 for Mind2Web. Most skills therefore consist primarily of a main document
and only a small number of attachments, which helps explain why file count is
generally a weak selector in Section~\ref{sec:nonllm}.

\section{Resource Provenance}
\label{sec:provenance}

The experimental materials referenced in this paper comprise code, data, and intermediate outputs organized in a \texttt{/Resources} directory. That directory is not included in this preprint's source package, and no public repository URL is provided in this version. The appendices contain the textual experimental details and scoring prompts; full computational reproduction additionally requires the separate experimental materials.